\documentclass[conference]{IEEEtran}
\IEEEoverridecommandlockouts
\usepackage{cite}
\usepackage{amsmath,amssymb,amsfonts}
\usepackage{algorithmic}
\usepackage{graphicx}
\usepackage{textcomp}
\usepackage{xcolor}
\usepackage{tabularx}
\usepackage{booktabs}
\usepackage{comment}
\usepackage{hyperref}
\usepackage{enumitem}
\usepackage[round, sort&compress, numbers]{natbib}
\begin{document}

\title{CU-UD: text-mining drug and chemical-protein interactions with ensembles of BERT-based models
}

\author{\IEEEauthorblockN{
Mehmet Efruz Karabulut$^1$ 
\hspace{4em}
K. Vijay-Shanker$^{1,*}$
\hspace{4em}
Yifan Peng$^{2,*}$}
\IEEEauthorblockA{\textit{$^1$Computer \& Information Sciences, University of Delaware, Newark, DE, USA} \\
\textit{$^2$Population Health Sciences, Weill Cornell Medicine, New York, NY, USA}}
\thanks{$^{*}$co-corresponding}
}

\maketitle


\begin{abstract}
Identifying the relations between chemicals and proteins is an important text mining task. BioCreative VII track 1 DrugProt task aims to promote the development and evaluation of systems that can automatically detect relations between chemical compounds/drugs and genes/proteins in PubMed abstracts. In this paper, we describe our submission, which is an ensemble system, including multiple BERT-based language models. We combine the outputs of individual models using majority voting and multilayer perceptron. Our system obtained 0.7708 in precision and 0.7770 in recall, for an F1 score of 0.7739, demonstrating the effectiveness of using ensembles of BERT-based language models for automatically detecting relations between chemicals and proteins. Our code is available at \url{https://github.com/bionlplab/drugprot\_bcvii}.
\end{abstract}


\section{Introduction}

Relation extraction is an important task in the biomedical domain that aims to identify the semantic relationship between the biomedical entities mentioned in the text. Information about these interactions can be helpful for other biomedical research such as drug discovery, database curation and question answering systems. In recent years, there have been lots of efforts to automatically extract the relations from the biomedical articles (e.g., \cite{peng2018extracting, zhang2018hybrid}). BioCreative events have focused on different relation extraction tasks over the years. One of the tasks in BioCreative VI was ChemProt \cite{krallinger2017overview}, which aims to promote the development and evaluation of systems that can automatically detect relations between chemical compounds and genes/proteins in PubMed abstracts. This task has been continued as a DrugProt task in BioCreative VII \cite{miranda2021overview}. In this paper, we discuss our participation in this task, describing ensemble approaches and incorporating multiple BERT-based language representation models.

\section{Methods}

\subsection{Dataset}

Organizers of the DrugProt track created a manually annotated corpus of 5,000 abstracts from PubMed \cite{miranda2021overview}. The dataset was split into a training set (3,500 abstracts), a development set (750), and a test set (750) (Table \ref{tab:dataset}). Following the settings of previous BioCreative tracks, the test set consists of a large collection of records containing a subset of a total of 750 Gold Standard records that was used for evaluation purposes \cite{krallinger2017overview}.

\begin{table}[]
    \centering
\caption{Statistics of the DrugProt dataset}
    \begin{tabular}{lrr}
\toprule
& Training & Development\\
\midrule
Documents & 3,500 & 750\\
Tokens & 1,001,168 & 199,620\\
Entities & 89,529 & 18,858\\	
\hspace{1em} Chemical & 46,274 & 9,853\\	
\hspace{1em} Gene & 43,255 & 9,005\\
Relations & 17,288 & 3,765\\
\hspace{1em} ACTIVATOR & 1,394 & 246\\
\hspace{1em} AGONIST & 652 & 131\\
\hspace{1em} AGONIST-ACTIVATOR & 27 & 10\\
\hspace{1em} AGONIST-INHIBITOR & 10 & 2\\
\hspace{1em} ANTAGONIST & 931 & 218\\
\hspace{1em} DIRECT-REGULATOR & 2,061 & 458\\
\hspace{1em} INDIRECT-DOWNREGULATOR & 1,297 & 332\\
\hspace{1em} INDIRECT-UPREGULATOR & 1,351 & 302\\
\hspace{1em} INHIBITOR & 5,277 & 1,152\\
\hspace{1em} PART-OF & 861 & 258\\
\hspace{1em} PRODUCT-OF & 904 & 158\\
\hspace{1em} SUBSTRATE & 1,921 & 495\\
\hspace{1em} SUBSTRATE\_PRODUCT-OF & 25 & 3\\
\bottomrule
    \end{tabular}
    \label{tab:dataset}
\end{table}

\subsection{Preprocessing}
Since all relations in the DrugProt corpus are enclosed in one sentence, our system considers interactions only occur within a sentence. We split the raw text into individual sentences by Stanza sentence splitter \cite{qi2022stanza}. These sentences are passed onto the BERT framework which tokenizes them using Wordpiece tokenizer \cite{wuwordpiece}.

Given a sentence with multiple entity mentions marked, we treated the relation extraction problem as a multi-class classification problem. Curators of the DrugProt corpus annotated thirteen key relations of biomedical importance. A sentence instance with a marked protein and a chemical mention is considered one of these thirteen DrugProt relation classes if there is a relationship between a protein and chemical occurrences. Otherwise, this instance is negative. There may be multiple interacting entities and non-interacting entities marked in the same sentence. We consider only a single chemical-protein pair in that sentence per sentence instance. Therefore, we formulate the task as predicting one of the fourteen relation types for a sentence instance.

We consider two entity tagging schemes (Table~\ref{tab:schemes}). In the first scheme, we anonymized target named entities in a sentence using pre-defined tags to prevent overfitting to entities \cite{peng2018transfer}. Specifically, given that each instance was comprised of a sentence and two entity mentions, we replaced each entity of interest with the tokens DRUG or PROTEIN, depending on what the specific mention represented. Other drugs and proteins are replaced with DRUG\_O and PROTEIN\_O, respectively. 

\begin{table}[]
    \centering
\caption{Examples of two entity tagging schemes}
\label{tab:schemes}
    \begin{tabular}{p{4em}p{24em}}
\toprule
Original sentence & human type 12 RDH reduces dihydrotestosterone to androstanediol\\
\midrule
Schema 1 & DRUG reduces dihydrotestosterone to PROTEIN\\
\midrule
Schema 2 & $<$DRUG-B$>$ human type 12 RDH $<$DRUG-E$>$ reduces dihydrotestosterone to $<$PROTEIN-B$>$ androstanediol $<$PROTEIN-E$>$\\
\bottomrule
    \end{tabular}
\end{table}

In the second scheme, we directly inserted the start and end entity markers into the original sentence. For example, we used the format $<$PROTEIN-B$>$ human type 12 RDH $<$PROTEIN-E$>$ to enclose the protein mention “human type 12 RDH” in the sentence.

\subsection{Model Development}

We hypothesize that an ensemble system that combines results of different models could lead to better predictive performance than using a single model \cite{peng2018extracting}. The superior performance of ensemble models is achieved when there is enough diversity among the individual models \cite{sagi2018ensemble}. In this study, we propose two different ensemble approaches which incorporate multiple BERT-based language representation models. Here, we chose BERT as the backbone because recent studies demonstrate it achieved state-of-the-art performance on a number of natural language understanding tasks, including relation extraction task \cite{devlin2018bert,su2021improving,peng2018transfer}.

We explore various architectures and pre-trained models to achieve the goal of diversifying BERT-based models. We will discuss these techniques and the ensemble algorithms in the following subsections. Due to the space limitations, we explain only the most relevant parts of the BERT model to the DrugProt task in the following. We refer the reader to the original paper for the details.

\vspace{1em}
\subsubsection{Single models} 
\begin{figure}[htbp]
\centering
\centerline{\includegraphics[width=.65\columnwidth]{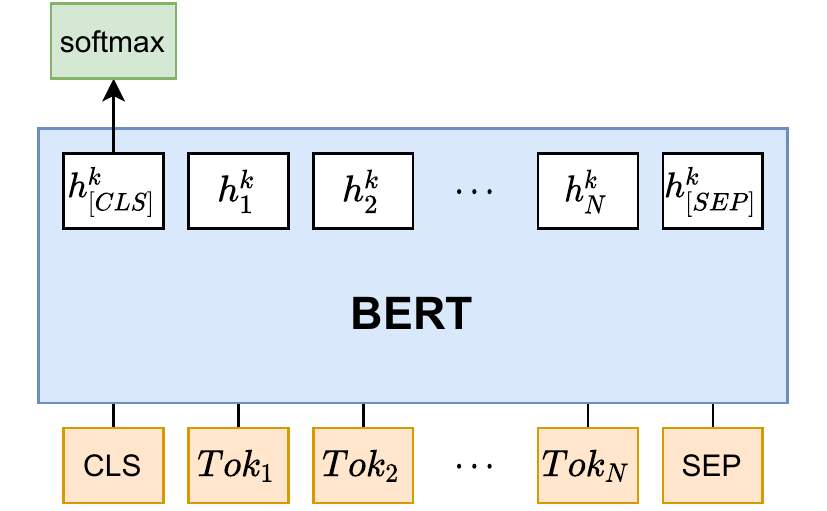}}
(a)\\[2em]
\centerline{\includegraphics[width=.65\columnwidth]{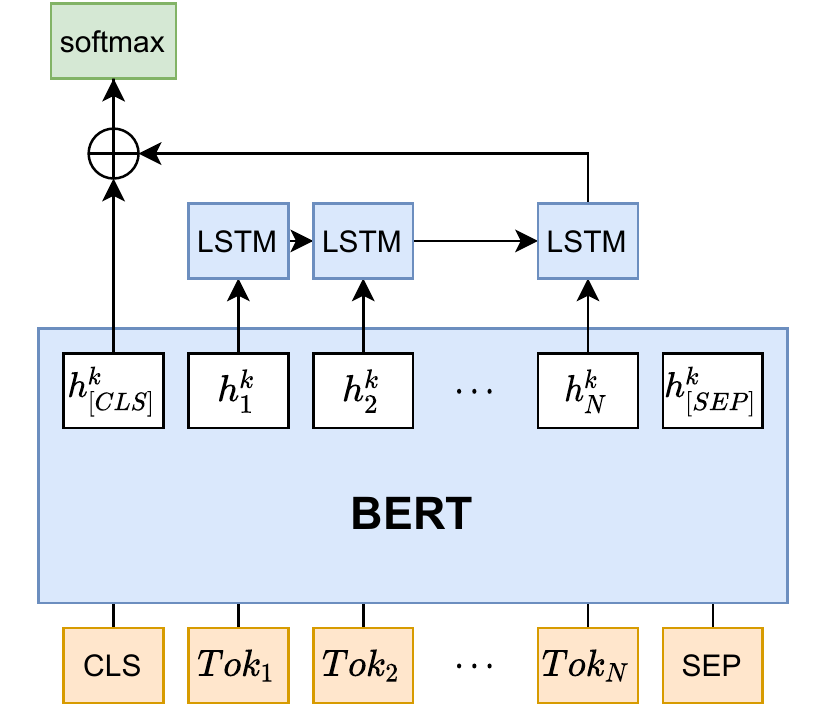}}
(b)\\[2em]
\centerline{\includegraphics[width=.65\columnwidth]{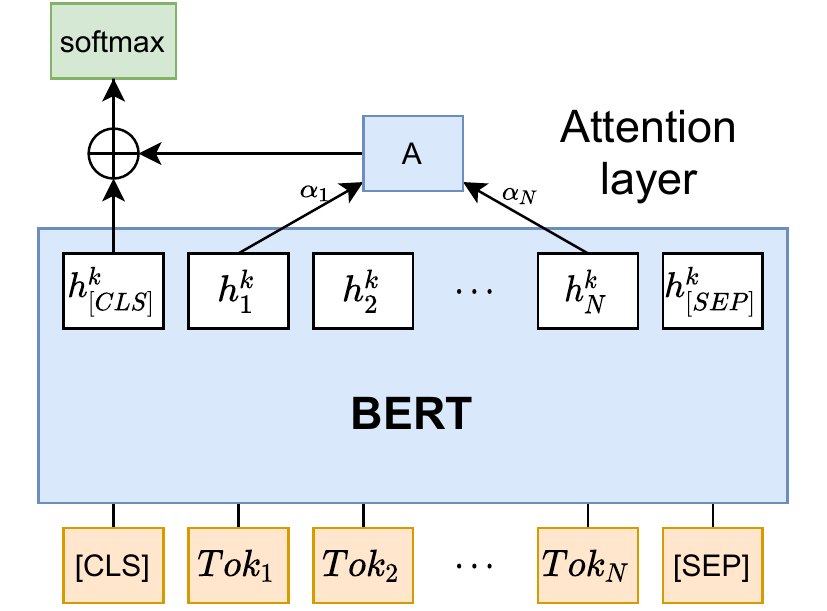}}
(c)
\caption{Architectures of individual models on relation extraction.}
\label{fig:architectures}
\end{figure}

We built three different architectures. The first one uses the last layer of the [CLS] token for the relation extraction task because it aggregates the information of the whole sentence. Fig.~\ref{fig:architectures}a demonstrates the fine-tuning process of BERT on the relation extraction task. A sentence instance is tokenized and embedded as in the original BERT paper. Inspired by the work of Su et al. \cite{su2020investigation}, we also explore two other methods of summarizing the information in the last layer: Long Short Term Memory (LSTM) on the last layer (Fig.~\ref{fig:architectures}b) and attention mechanism on all units of the last layer (Fig.~\ref{fig:architectures}c).

To obtain diversity among the models used in our ensembles, we chose three BERT-based models in our approach, BioBERT \cite{lee2020biobert}, PubMedBERT \cite{gu2020domain}, and BioM-ELECTRA  \cite{alrowili2021biom,clark2020electra}, since they differ in some pre-training aspects as well as masking, vocabulary, and the learning objectives techniques. The pre-training of BioBERT used  PubMed and PMC after the initial pre-training of BERT, whereas PubMedBERT pre-training using PubMed alone started from scratch. These two models are BERT-Base with 12 bidirectional self-attention heads. To study the effectiveness of large models for this task, we also used BioM-ELECTRAL. BioM-ELECTRAL is pre-trained on PubMed abstracts with no previous pre-training and is based on the ELECTRA architecture \cite{alrowili2021biom}. 

\vspace{1em}
\subsubsection{Class-weighted loss function}
Since we have created a negative instance for each non-interacted chemical and protein in the same sentence, we have substantially more negative examples in the training dataset. To mitigate this unbalance, we assigned class weights inversely proportional to their respective frequencies. 

\vspace{1em}
\subsubsection{Majority voting}
Majority voting is the simplest yet effective weighting method for output fusion. We select the relation type that gets the most votes out of five individual models. If the models cannot agree on the majority vote, we assign this instance as negative.

\vspace{1em}
\subsubsection{Meta-learning method}
\label{sec:mlp}
\begin{figure}
\centering
\includegraphics[width=.75\columnwidth]{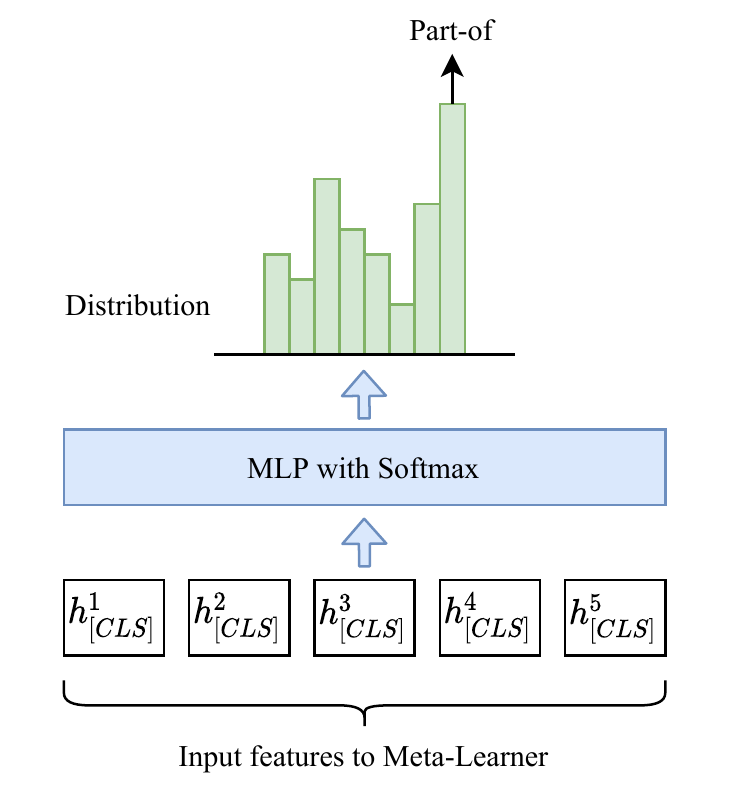}
\caption{The architecture of the MLP-based ensemble model. $h^i_{[CLS]}$ indicates the [CLS] token output of the final layer for the $i$’th model in the ensemble.}
\label{fig:mlp}
\end{figure}

We also employ another ensemble method using multilayer perceptron (MLP) as a meta-learner to learn from individual classifiers. Specifically, we use MLP to combine the outputs of each individual model in an ensemble (Fig. \ref{fig:mlp}). After each individual model is fine-tuned, we extract the last output layer of the [CLS] token from each model. We then use the development set that we reserve for the ensemble learning to obtain the activation vector for the [CLS] token. Afterward, we concatenate them into one vector and pass it to another MLP. We insert a softmax layer on top of the MLP to obtain final predictions.

\section{Results and Discussion}

\subsection{Experimental settings}

We submitted five runs. For Runs 1 and 2, we used the five individual models listed in Group A (PubMedBERT, PubMedBERT-CW, PubMedBERT-LSTM, BioBERT-Attention,  Biobert-Tag2) and Group B (PubMedBERT, PubMedBERT-CW, PubMedBERT-LSTM, BioM-ELECTRA$_{L}$,  BioBERT-Tag2). “Tag2” denotes the second entity tagging scheme. ``CW" indicates that this model was fine-tuned using class-weighted loss function.


Individual model predictions were combined using a stacking approach involving a MLP with one hidden layer having 512 units as described in Section \ref{sec:mlp}. Run 3 involves the same Group B and uses both stacking and the majority voting. Runs 4 and 5 apply only one type of model, BioM-ELECTRA$_{L}$. Five models are obtained for Run 4 by training on five different partitions. However, Run 5 involves only one model, allowing us to see the extent of the gain by using an ensemble method.   

While participant teams were provided with the training data set and the development data set, we combined them together to develop the models.  For Runs 1 and 2, we first used 70\% of the total data for training, 20\% for development (tuning the hyper-parameters for the individual models) and 10\% to train the ensemble models. For the final models, we trained the individual models on the original training data and ensemble models on the development data with the chosen hyper-parameters.

\begin{figure}
\centering
\includegraphics[width=.8\columnwidth]{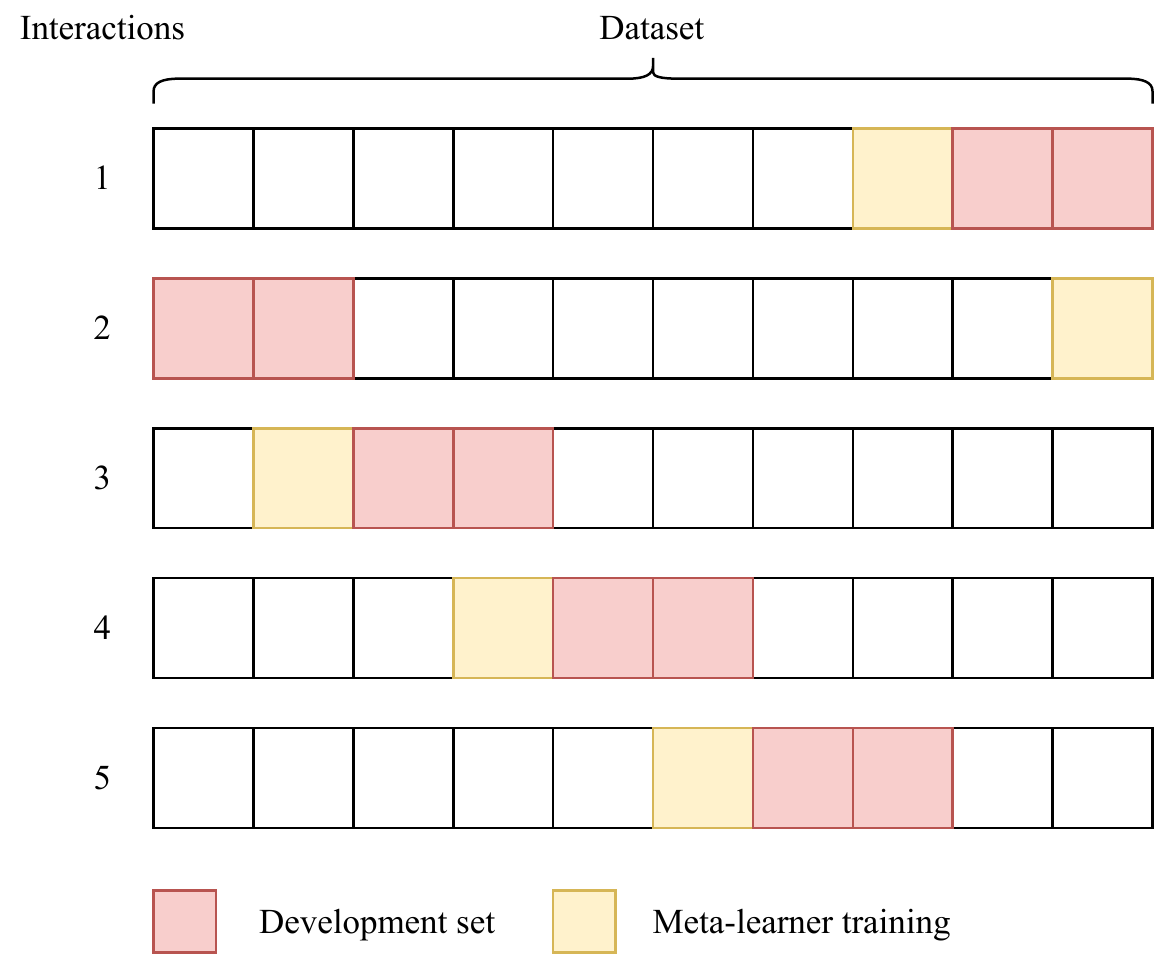}
\caption{70-20-10 split is performed 5 times.}
\label{fig:5cv}
\end{figure}

For Run 3, we used the same approach of 70-20-10 split, but repeated this process five times (Fig \ref{fig:5cv}). Thus, we obtained a group of five models corresponding to each 70-20-10 partition. Next, we trained a separate MLP for each group of models using the 10\% as described above. As a result, we have five predictions for each instance corresponding to each MLP. Finally, we combined them using majority voting to obtain the final predictions. 

For Run 4, we fine-tuned five BioM-ELECTRA$_{L}$ models using 80\% of total data in each fold and then combined their predictions using majority voting. 

Run 5 involved only one model, and hence 100\% of the data was used for its fine-tuning.

Our models are implemented in Tensorflow 1.14 and all the experiments are carried out on NVIDIA 2080Ti GPUs. We employ AdamW optimizer with an initial learning rate of 2e-5. The maximum epoch is set as 10. 

\subsection{Results and discussions}

Table \ref{tab:resultsval}
and \ref{tab:results} shows the results we obtained for 5 Runs on the development set in iteration 1 (Fig~\ref{fig:5cv}) and the official test set, respectively. 
Runs 3 and 4 have the best F1 performance, which is substantially better than the average F1 across participant teams. Interestingly, F1 score of the single large model is almost the same as those of the ensembles of Runs 1 and 2. The best F1 scores of Runs 3 and 4 are due to the increase of Precision even if the Recall dropped slightly.

\begin{table}[]
    \centering
    \caption{Results of our systems on the development set in iteration 1 (Fig~\ref{fig:5cv})}
    \label{tab:resultsval}
    \begin{tabular}{clccc}
\toprule
Run & System & P & R & F1\\
\midrule
1 & Stacking (MLP) & 0.7630 & 0.7720 & 0.7674\\
2 & Stacking (MLP) & 0.7700 & 0.7838 & 0.7764\\
3 & Stacking (MLP)+Majority Voting & \textbf{0.7770} & \textbf{0.7780} & \textbf{0.7770}\\
4 & Majority Voting & 0.7622 & 0.7832 & 0.7726\\
5 & BioM-ELECTRA$_{L}$ & - & - & -\\

\bottomrule
    \end{tabular}
\end{table}


\begin{table}[]
    \centering
    \caption{Results of our systems on DrugProt test set}
    \label{tab:results}
    \begin{tabular}{clccc}
\toprule
Run & System & P & R & F1\\
\midrule
1 & Stacking (MLP) & 0.7421 & 0.7902 & 0.7654\\
2 & Stacking (MLP) & 0.7360 & 0.7925 & 0.7632\\
3 & Stacking (MLP)+Majority Voting & \textbf{0.7708} & \textbf{0.7770} & \textbf{0.7739}\\
4 & Majority Voting & 0.7721 & 0.7750 & 0.7736\\
5 & BioM-ELECTRA$_{L}$ & 0.7548 & 0.7747 & 0.7647\\
\midrule
& Mean of participant teams & 0.6430 & 0.6291 & 0.6196\\
\bottomrule
    \end{tabular}
\end{table}

Table \ref{tab:results2} shows the results of individual relation types using our best system (Run 3). Our system obtained the best performance of 0.8902 in F1 for the ANTAGONIST relation, and 0.8570 for the INHIBITOR relation. 
As shown in Table \ref{tab:dataset}, INHIBITOR is the most frequent relation in the dataset. Thus its classification is probably well trained. On the other hand, ANTAGONIST is one of the least frequent relations. Further error analysis is needed to understand why our model can effectively extract it with such few cases. Compared to results of individual relation types by other Runs, we observed that Runs 3 and 4 have a uniform increase in F1 performance for almost all individual relation types. On the other hand, Run 1 has the best F1 performance on the AGONIST type with a 2.5\% increase over Run 3. 

\begin{table}[]
    \centering
    \caption{Results of individual relation types using our best system}
    \label{tab:results2}
    \begin{tabular}{lccc}
\toprule
Relation & P & R & F1\\
\midrule
ACTIVATOR & 0.8083 & 0.8083 & 0.8083\\
AGONIST & 0.7857 & 0.7623 & 0.7738\\
ANTAGONIST & 0.8342 & 0.9542 & 0.8902\\
DIRECT-REGULATOR & 0.7579 & 0.6643 & 0.7080\\
INDIRECT-DOWNREGULATOR & 0.6976 & 0.7894 & 0.7407\\
INDIRECT-UPREGULATOR & 0.7260 & 0.7942 & 0.7586\\
INHIBITOR & 0.8586 & 0.8553 & 0.8570\\
PART-OF & 0.6781 & 0.7763 & 0.7239\\
PRODUCT-OF & 0.6285 & 0.7292 & 0.6751\\
SUBSTRATE & 0.7189 & 0.6348 & 0.6742 \\
\bottomrule
    \end{tabular}
\end{table}

\section{Conclusion}

In this manuscript, we describe our submission in the 
BioCreative VII DrugProt task. The results demonstrate that 
our ensemble system can effectively detect the chemical-protein relations from biomedical literature.

\section*{Acknowledgment}

This work is supported by the National Library of Medicine under Award No. 4R00LM013001.

\footnotesize{

}

\end{document}